\documentclass{article}
\usepackage{spconf,amsmath,graphicx}
\usepackage{multirow}
\usepackage{amsmath}

\title{Learning Using Privileged information for segmenting tumors on digital mammograms}
\name{Ioannis N. Tzortzis$^1$, Konstantinos Makantasis$^2$, Ioannis Rallis$^1$, Nikolaos Bakalos$^1$, \thanks{This work has been supported by the European Union's Horizon 2020 research and innovation programme from the INCISIVE project (Grant Agreement No. 952179) and the H2020 SECURED project, grant agreement No. 101095717.}}
\address {1: National Technical University of Athens, Computer Vision and Photogrammetry Laboratory \\
2: University of Malta, Department of Artificial Intelligence}

\begin{document}
%\ninept
%
\maketitle
\begin{abstract}
Limited amount of data and data sharing restrictions, due to GDPR compliance, constitute two common factors leading to reduced availability and accessibility when referring to medical data. To tackle these issues, we introduce the technique of Learning Using Privileged Information. Aiming to substantiate the idea, we attempt to  build a robust model that improves the segmentation quality of tumors on digital mammograms, by gaining privileged information knowledge during the training procedure. Towards this direction, a baseline model, called student, is trained on patches extracted from the original mammograms, while an auxiliary model with the same architecture, called teacher, is trained on the corresponding enhanced patches accessing, in this way, privileged information. We repeat the student training procedure by providing the assistance of the teacher model this time. According to the experimental results, it seems that the proposed methodology performs better in the most of the cases and it can achieve 10\% higher F1 score in comparison with the baseline.
\end{abstract}
\section{Introduction}
\label{sec:intro}

Breast cancer appears to be the leading type of cancer in females, from the perspective of incidence worldwide (except from Eastern Africa, Oceania and Australia/New Zealand where breast cancer comes second on the list), according to cancer statistics for the year 2020 \cite{ferlay2021cancer}. With regard to the mortality rate, breast cancer shares the leading position with Cervix uteri and Lung cancer, accounting almost 685.000 deaths in 2020 worldwide \cite{tzortzis2022tensor}. The recent research \cite{siegel2023cancer}, published in 2023, shows that breast cancer still remains first on the list of estimated new cases and second on the list of estimated deaths in females considering the United States of America.

As stated in \cite{liu2023mammography}, mammography is the most commonly used technique for the early detection of breast cancer, mainly due to its reproducible results and its relative low cost. Undoubtedly, it is suggested that women over the age of 40 should undergo annual mammography examination \cite{bevers2023nccn}. Even though mammography is the most common method used for the early detection of the breast cancer, it seems to be critisized a lot in consequence to 2 major reasons: a) its high false-positives rate and b) the hyper-diagnosis of low-risk diseases \cite{kuhl2023future}.

Through the wide utilization of mammography at the primary stages of the clinical workflow, plethora of Artificial Intelligence tools have been developed in an attempt to improve the early detection of the cancer and assist the medical professionals during the examination process \cite{abhisheka2023comprehensive}. Additionally, plenty of more advanced techniques like transfer learning, generative networks, knowledge distillation, synthetic data have been introduced aiming to improve the classification/detection/segmentation/augmentation tasks and extract more useful information from mammograms\cite{zhang2023recent}. With such an attention to these tasks, the corresponding techniques have achieved high performance.

Thus, the upcoming new challenge is related to the availability of data. The existing publicly available datasets, containing mammograms, are limited and, in many case, appear to be insufficient on concerning specific characteristics like the dataset volume, the quality of data, the precision of related information like annotation masks, etc \cite{logan2023review}. Additionally, due to GDPR (General Data Protection Regulation) restrictions, it is difficult and time consuming to utilize available data exist in medical centers since it requires extensive bureaucratic procedures.

Heading towards the direction of tackling these issues, we propose the utilization of LUPI (Learning Using Privileged Information) \cite{vapnik2015learning} technique in order to train a superior student model for the accurate segmentation of tumors on mammograms. The novelties of the proposed model can be summarized in the following points. Firstly, the privileged information model perform well even with low amount of training data and appears to be more robust. Additionally, the generalization of the model tends to be more inclusive. The privileged information should be obtained only during the training process, meaning that the model acquires solely the mammogram during the testing. Furthermore, the proposed methodology can assist other existing techniques without changing any part of the architecture; i.e. a federated learning schema could be enhanced by using out proposed model for achieving faster and more robust distributed learning. 

\subsection{Related work}
\textbf{Multi-modal networks: }The authors in \cite{habib2020automatic}, present a multi-modal model for the classification of breast lesions. The final model anticipates receiving jointly features produced by two single-modal models.
In \cite{song2021multiview}, a multiview multimodal network is introduced that incorporates CC (craniocaudal) and MLO (mediolateral oblique) view for both LE (Low-energy) and DES (Dual-energy subtracted) images.
Both of the aforementioned techniques merge different modalities together in order to enhance the training of the models. Though, the additional modalities should be used during the testing procedure as well.

\noindent\textbf{Knowledge distillation schemes: }A knowledge distillation scheme is presented in \cite{sepahvand2023joint} for the classification of histopathological images. The authors attempt to develop a lightweight learning model that permorms as well as the state-of-the-art models. 
The authors in \cite{kanwal2023vision} try to boost the performance of a Vision Transformer  model, developed for detecting air bubbles on histological images, by distilling knowledge from a high-capacity teacher model. The main drawback of these schemes is the increased complexity because of the different model architectures selected for the teache models. 

\noindent\textbf{Transfer learning approaches:} A transfer learning, using privileged information, schema is presented in \cite{shaikh2020transfer} for the classification of mammograms. Ultrasound images are utilized as the major modality, while corresponding mammograms are considered privileged information.
The authors in \cite{fei2021doubly}, propose a LUPI based transfer learning to enhance the BUS (bimodal ultrasound) CAD (computer aided diagnosis) by utilizing information from EUS (elastography ultrasound) modality. These techniques follow common principles in comparison with the proposed approach. Though, they aim towards the classification tasks and their main modality is the ultrasound images.

\begin{figure}[ht]
\includegraphics[width=\linewidth]{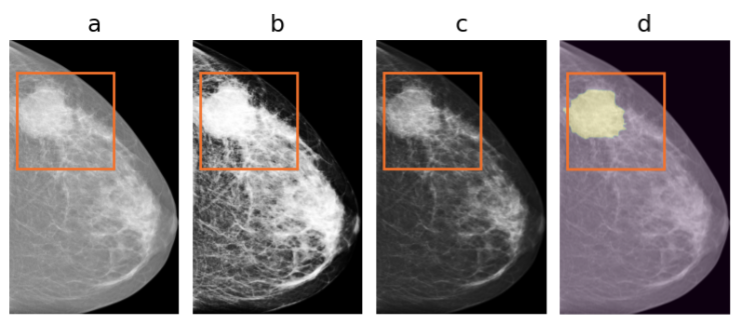}
\caption{a) The original mammogram, b) the output after applying histogram equalization, c) the output after applying contrast adjustment, d) the corresponding annotation mask.}
\label{fig: filters}
\centering
\end{figure}

\section{Methodology}
\label{sec:method}

\subsection{Problem formulation}
Our collection consists of square patches extracted from the original mammograms. 
It is clear we are dealing with a semantic segmentation problem, which, assuming the pixel wise processing, boils down to a binary classification task. The first class represents the healthy tissue, while the second one corresponds to the non healthy tissue which contains a tumor or part of it. 

To conduct our experiments, it is essential to define a baseline model in order to make the proper comparisons. This model will be the student case itself, meaning that we will train the model solely with the mammogram patches.
To formulate this, let $I_i \in \mathcal X_s$ denotes the information provided by the $i_{th}$ patch and $Y_i$ the corresponding annotation mask which contains pixel-wise ground truth information; pixels with value equal to zero represent the healthy tissue, while pixels with value equal to zero show the tumorous tissue. Since the collection of such pairs refer to the student model, we call it $D_s$ (Eq. \ref{eq:dataset}).

\begin{equation}
\label{eq:dataset}
    \mathcal D_s = \{(I_i, Y_i)\}_{i=1}^N
\end{equation}

\noindent Let $x$ represent a random sample from the collection $D_s$ that is fed to the student model during the training process. Then, the student model is minimizing the $L_{st}$ loss, described in Eq. \ref{eq:lst}.
\begin{equation}
\label{eq:lst}
    L_{st} = L_{CE}(S_o(x), y)
\end{equation}

\noindent where $L_{CE}$ stands for the Cross-Entropy loss function, $S_o(x)$ denotes the output of the student model on the given input and $y$ represents the ground truth labels of the selected sample $x$.

For the teacher case, we employ the same model architecture with different input channels, since we have enhanced the patches with the outputs of specific filters. So, let $\Bar{I_i} = [I_i^1 I_i^2 I_i^3] \in \Bar{X_t}$ denotes the information provided by the enhanced $i_{th}$ patch and $Y_i$ the corresponding annotation mask. $I_i^1$ is the same image as the original one used by the student as input ($I_i$), $I_i^2$ refers to the image occur from the histogram equalization applied on the original patch and $I_i^3$ represents the output of the contrast adjustment filter applied on the original patch. $Y_i$ refers to the same annotation mask of the $i_{th}$ patch as the student model (Figure \ref{fig: filters}). The collection of such pairs ($\Bar{I_i}, Y_i$) constitute the dataset $D_t$ (Eq. \ref{eq:t_dataset}) employed for the training of the teacher model. 

\begin{equation}
\label{eq:t_dataset}
    \mathcal D_t = \{(\Bar{X_i}, l_i)\}_{i=1}^N
\end{equation}

Finally, considering the privileged information case, the student not only attempts to adapt its probabilistic predictions to the ground truth but, at the same time, it tries to fit them to the corresponding teacher's probabilistic predictions. In a nutshell, the student is minimizing the loss $L_{PI}$, shown in Eq. \ref{eq:tsloss}, during the training process.

\begin{multline}\label{eq:tsloss}
    L_{PI} = \alpha * L_{CE}(S_o(x), y) +\\
    (1 - \alpha) *  L_{CE}(S_o(x), T_o(\Bar{x}))
\end{multline}

where $L_{CE}$ is the Cross-Entropy loss, $S_o(x)$ denotes the probabilistic predictions of the student model given a sample batch $x$ of images from the $D_s$, $T_o(\Bar{x})$ represents the probabilistic predictions of the teacher model given a sample batch $\Bar{x}$ of images from the $D_t$ and $a$ is a scalar that takes values in the interval $[0, 1]$. The parameter $\alpha$ defines the affect level of the teacher on the student. The notation used in this section is borrowed by the paper \cite{makantasis2023lab}.

\subsection{The segmentation model architecture}

\begin{figure}[ht]
\label{fig:unet}
\includegraphics[width=\linewidth]{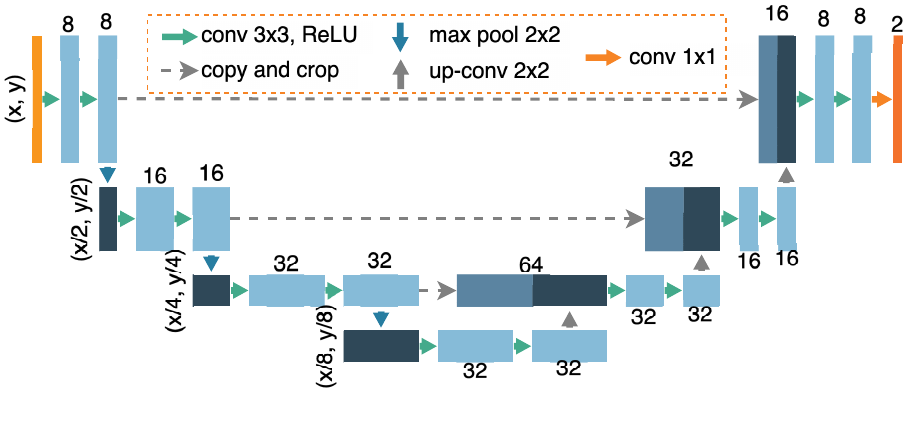}
\caption{The UNet model}
\centering
\end{figure}
The deep learning model, used for the purposes of this work, is the UNet model presented in Figure \ref{fig:unet}. It is based on the architecture appeared in \cite{ronneberger2015u}, which was proposed for biomedical image segmenation purposes. It consists of three sequential contracting blocks that are followed by three sequential expanding blocks. Both the contracting and the expanding blocks include 2 sequences of a convolution layer, a batch normalization operation and a ReLU activation function. The former leads to a max pooling operation, while the latter to transposed convolutional layer. 

% \subsection{Privileged information}

\section{Experimental results}
\label{sec:exp}

\begin{table*}[ht]

\centering
\begin{tabular} { |c|c|c|c||c|c|c|c|c|  }
 \hline
 \multicolumn{9}{|c|}{Experimentation map} \\
 \hline
 Exp id & TF & Range & Iters& Teacher & Student & PI - $a=0.8$ & PI - $a=0.6$ & PI - $a=0.4$\\
 \hline
 E1 & 1 & 1-400 &  \multirow{16}{1em}{5}   & 0.654 $\pm$ 0.06 & 0.54 $\pm$ 0.26 & 0.565 $\pm$ 0.13 & \textbf{0.631 $\pm$ 0.05} & 0.608 $\pm$ 0.05\\ 
 E2 & 1 & 1-600 &    & 0.559 $\pm$ 0.09 & 0.436 $\pm$ 0.13 & \textbf{0.564 $\pm$ 0.1} & 0.467 $\pm$ 0.33 & 0.405 $\pm$ 0.29\\
 % \hline
 E3 & 1 & 1-800 &    & 0.655 $\pm$ 0.05 & 0.56 $\pm$ 0.12 & \textbf{0.646 $\pm$ 0.02} & 0.646 $\pm$ 0.04 & 0.643 $\pm$ 0.04\\
 % \hline
 E4 & 1 & 1-1000 &    & 0.621 $\pm$ 0.18 & 0.552 $\pm$ 0.17 & 0.531 $\pm$ 0.13 & \textbf{0.65 $\pm$ 0.06} & 0.611 $\pm$ 0.12\\
 % \hline
 E5 & 2 & 1-400 &    & 0.621 $\pm$ 0.08 & 0.607 $\pm$ 0.08 & 0.562 $\pm$ 0.18 & 0.63 $\pm$ 0.04 & \textbf{0.636 $\pm$ 0.04}\\
 % \hline
 E6 & 2 & 1-600 &    & 0.541 $\pm$ 0.22 & 0.604 $\pm$ 0.07 & 0.54 $\pm$ 0.12 & \textbf{0.606 $\pm$ 0.03} & 0.525 $\pm$ 0.17\\
 % \hline
 E7 & 2 & 1-800 &    & 0.652 $\pm$ 0.06 & 0.603 $\pm$ 0.11 & 0.614 $\pm$ 0.08 & 0.638 $\pm$ 0.07 & \textbf{0.646 $\pm$ 0.03}\\
 % \hline
 E8 & 2 & 1-1000 &    & 0.555 $\pm$ 0.12 & 0.59 $\pm$ 0.11 & 0.534 $\pm$ 0.07 & \textbf{0.615 $\pm$ 0.03} & 0.592 $\pm$ 0.04\\
% \hline
 E9 & 3 & 1-400 &   & 0.666 $\pm$ 0.04 & 0.598 $\pm$ 0.1 & 0.462 $\pm$ 0.13 & 0.619 $\pm$ 0.11 & \textbf{0.676 $\pm$ 0.03}\\
% \hline
 E10 & 3 & 1-600 &    & 0.617 $\pm$ 0.04 & \textbf{0.646 $\pm$ 0.05} & 0.523 $\pm$ 0.21 & 0.594 $\pm$ 0.04 & 0.621 $\pm$ 0.03\\
% \hline
 E11 & 3 & 1-800 &    & 0.665 $\pm$ 0.05 & 0.526 $\pm$ 0.18 & \textbf{0.656 $\pm$ 0.03} & 0.63 $\pm$ 0.06 & 0.628 $\pm$ 0.07\\
% \hline
 E12 & 3 & 1-1000 &    & 0.618 $\pm$ 0.1 & 0.584 $\pm$ 0.1 & 0.499 $\pm$ 0.12 & \textbf{0.656 $\pm$ 0.03} & 0.618 $\pm$ 0.09\\
% \hline
 E13 & 4 & 1-400 &    & 0.637 $\pm$ 0.03 & 0.559 $\pm$ 0.13 & 0.642 $\pm$ 0.06 & 0.651 $\pm$ 0.04 & \textbf{0.657 $\pm$ 0.03}\\
% \hline
 E14 & 4 & 1-600 &    & 0.669 $\pm$ 0.04 & 0.502 $\pm$ 0.27 & 0.622 $\pm$ 0.04 & 0.641 $\pm$ 0.02 & \textbf{0.655 $\pm$ 0.06}\\
% \hline
 E15 & 4 & 1-800 &    & 0.659 $\pm$ 0.05 & 0.558 $\pm$ 0.13 & 0.583 $\pm$ 0.05 & \textbf{0.671 $\pm$ 0.04} & 0.662 $\pm$ 0.05\\
% \hline
 E16 & 4 & 1-1000 &    & 0.668 $\pm$ 0.05 & 0.586 $\pm$ 0.14 & 0.551 $\pm$ 0.13 & \textbf{0.652 $\pm$ 0.02} & 0.623 $\pm$ 0.03\\
 \hline
\end{tabular}
\caption{The experimentation map. For each experiment, we report the general details (experiment id, training fold, image range, iters) and the corresponding performance of the models followed by the 95\% confidence intervals. With bold-weighted font, we highlight the model with the best performance among the baseline/student model and the three case of the proposed one.}
\label{table: results}
\end{table*}

\begin{figure*}[ht]
\label{fig:bars}
\centering
\includegraphics[width=0.49\linewidth]{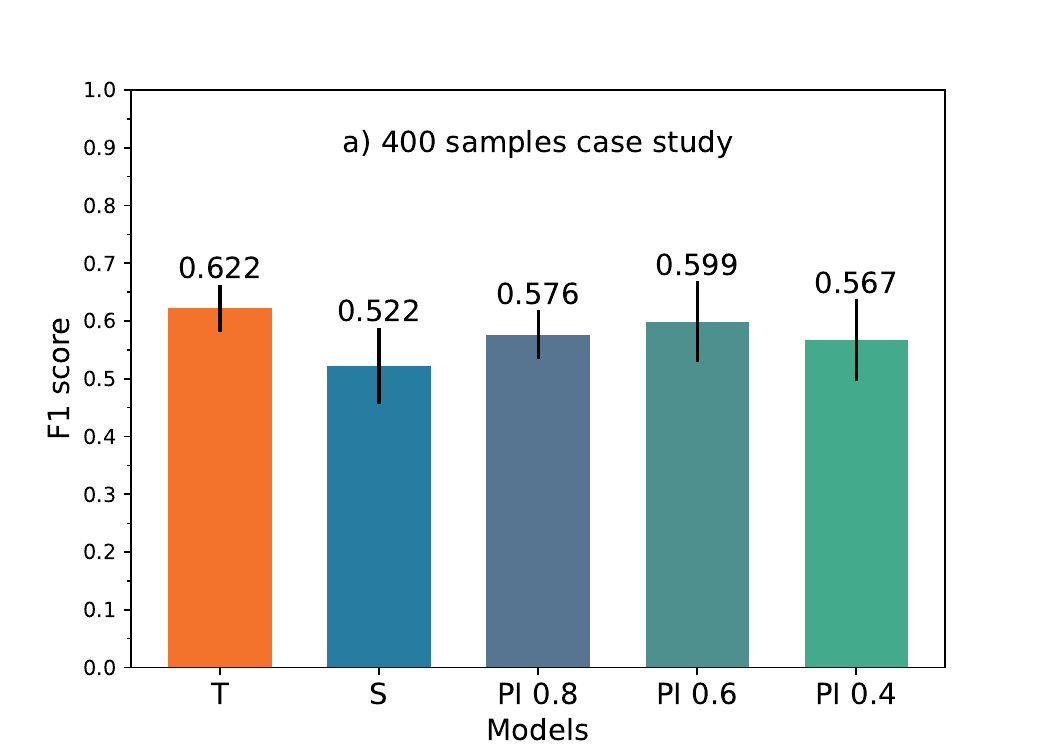}
\includegraphics[width=0.49\linewidth]{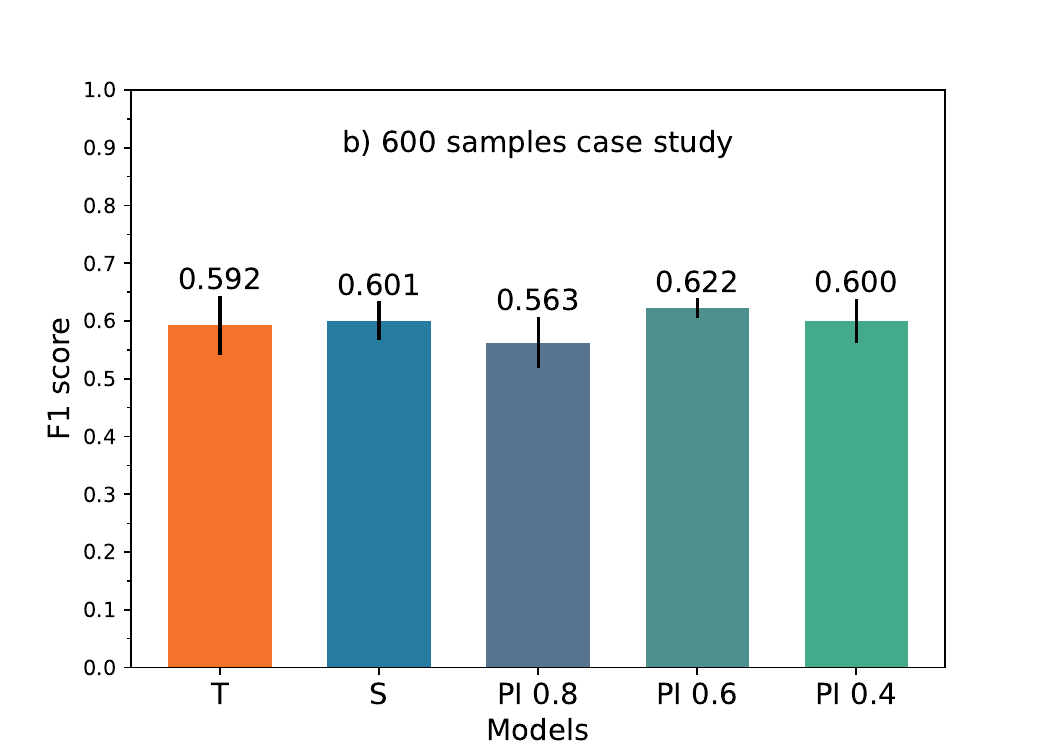}
\includegraphics[width=0.49\linewidth]{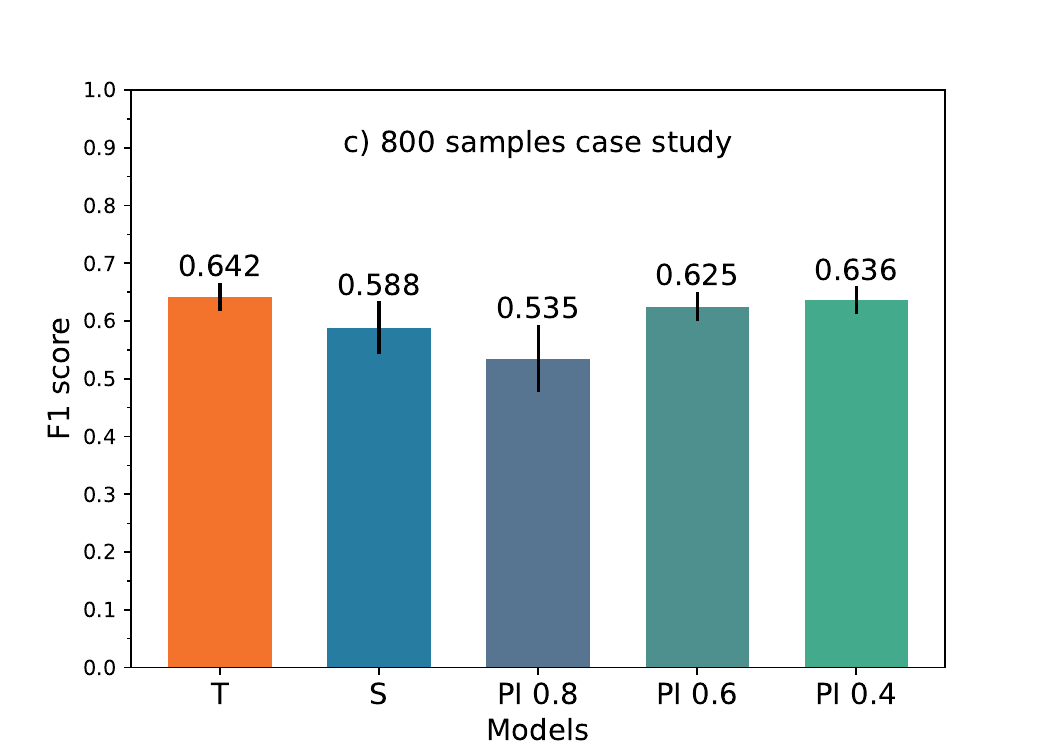}
\includegraphics[width=0.49\linewidth]{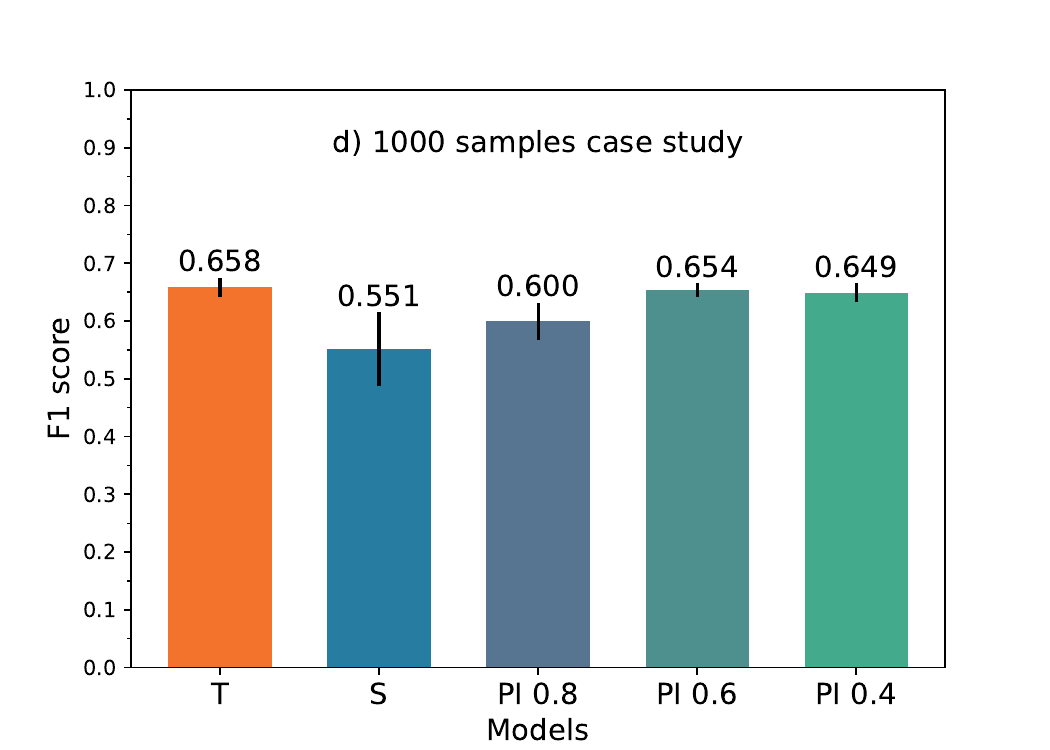}
\caption{The experimental results grouped by the amount of samples utilized for the training of the models. Each plot shows the performance of all the models which is measured using the F1 score.}

\end{figure*}

\subsection{Dataset description}

The original dataset, utilized for the purposes of this paper, is the INBreast digital mammographic database \cite{moreira2012inbreast}. It contains 410 digital mammograms that correspond to 115 patients. 90 of them have both breasts affected by abnormalities (tumors, calcifications, etc.), which are depicted in 4 mammograms. 25 patients have undegone mastectomy and, thus, these cases include 2 mammograms per case. For each mammogram, a matching annotation mask exists containing countour information about the several types of lesions.

For the objectives of this work, we considered each mammogram as a standalone image and did not take into account the relation among them. In order to acquire an essential data collection, we augmented the INBreast dataset by extracting patches from the original mammograms. In our case, we utilize only the mammograms containing tumors. Towards this direction, we are looking for the mammograms that correspond to non empty annotation masks. 

\subsection{Data pre-processing}

\begin{table}[ht] 
\centering
\begin{tabular} { |c||c|c|  }
 \hline
 \multicolumn{3}{|c|}{Experimentation map} \\
 \hline
 Parameter & Description & Value \\
 \hline
 $h_{ppi}$ & Healthy patches per image & 40 \\ \hline
 $nh_{ppi}$ & Non-healthy patches per image & 40 \\ \hline
 $mar$ & Mass area ration threshold & 0.01 \\ \hline
 $bar$ & Mass area ration threshold & 0.8 \\ \hline

\end{tabular}
\caption{The parameters values for the patch extraction procedure.}
\label{table:params}
\end{table}
 
For each of the selected images, our software attempts to locate, randomly, $h_{ppi}$ and $nh_{ppi}$ patches of size 1024x1024. $h_{ppi}$ stands for healthy-patches-per-image and denotes the number of healthy patches that will be extracted from the image. $nh_{ppi}$ (non-healthy patches per image) represent the number of non-healthy patches that will be extracted from the image. The validity of an extracted patch depends on two major properties; a) the mass-area-ratio ($mar$) and b) the breast-area-ratio ($bar$). The former property indicates the least area that should be covered by tumorous tissue within the patch. Accordingly, the latter property denotes the minimum surface that should be covered by breast tissue (healthy or not) in order for the patch to be considered as valid. The exact values of the parameters are shown in Table \ref{table:params}. An example of a valid patch is annotated with orange color in Figure \ref{fig: filters}.

Following the aforementioned requirements, we created a well-balanced collection of 5628 patches, which correspond to 108 patients with mammograms that contain tumors. Before the creation of the patches, we decided to use 88 patients for the training procedure and 20 for the testing of the models. The 88 patients refer to the first 4500 patches, while the rest 20 refer to the last 1128. Hence, we ensure that there is no case were a patch from the training set and a patch from the testing set belong to the same patient.

\subsection{Experimental setup}
In order to set a series of experiments, we initially split the 4500 training patches into 4 equally sized folds (1125 patches per fold). For each training fold, we introduce 4 different experiments with different amount of samples for the training procedure (400, 600, 800 and 1000 samples). Each experiment includes the teacher model training, the student model training and three cases of the proposed privileged information training using three different values of the parameter $\alpha$. We conducted each experiment 5 times in order to ensure the convergence of the results and calculate potential variance. For the training procedure of all models, we employed a 5-fold cross-validation schema to confirm the convergence of the results within a specific dataset region.

\subsection{Performance evaluation}

More tangible details about the experiments are presented in Table \ref{table: results}. The columns shows the a) experiment id, b) the selected training fold, c) the image range within the fold, d) the iterations the experiment conducted, e) the teacher model results, f) the student/baseline model results, g) the Privileged Information model results when $\alpha=0.8$, h) the Privileged Information model results when $\alpha=0.6$ and the Privileged Information model results when $\alpha=0.4$. Note that the quantity $1-\alpha$ shows the affect of the teacher model on the student training. When $\alpha=0$ the privileged information model becomes the same with the baseline/student model. All the results present the 95\% confidence interval that occur from the several repetitions of an experiment. 

Except from the experiment E10, it is clear that the  proposed model presents higher F1-scores at on of the three different versions ($\alpha=0.8$, $\alpha=0.6$ or  $\alpha=0.4$). In all these cases, it is remarkable the fact that the proposed model present lower variance in comparison with the baseline. Aiming to have a more clear view concerning the performance of the models, we group together the experiments that refer to the same image range, creating the bar plots of Figure \ref{fig:bars}. The upper left plot presents the scores achieved by the models when tested on 400 samples regardless the training fold that they belong. In other words, we combined the results from experiments E1, E5, E9 and E13. Accordingly, the same principle was adopted for the cases of 600, 800 and 1000 samples. According to the bar plots we can observe the following:

\begin{itemize}
    \item in all cases, at least one of the 3 proposed model versions is superior in comparison with the student/baseline model, 
    \item except from the 600 samples case, the privileged information with $\alpha=0.6$ seems to be the most proper configuration,
    \item the proposed model can achieve almost 10\% higher performance than the baseline. The lowest improvement equals to 2.1\% and refers to the 600 samples case,
    \item except from the 400 samples case, the proposed model presents a more robust behaviour by observing the variance of the results. Even in the first case, the privileged information model with $\alpha=0.8$ does not present the best F1 score but provides better variance.
\end{itemize}

\section{Conclusion}
In this paper, we employed the idea of Learning Using Privileged Information in order to tackle the issues of limited amount of medical data and the restrictions for data sharing from the hospitals side due to the compliance to the GDPR. To demonstrate the proof of our concept, we focused on the breast tumor segmentation techniques applied on patches extracted from digital mammograms. According to the experimental results, it seems that the proposed model outperforms the baseline model and can increase the F1-score even for 10\%. Additionally, the proposed architecture tends to be more robust, concerning the decreased variance observed in the results. For sure, the principles of the technique appear to solid showing the potentials to several applications. Employing more datasets or trying different privileged information could be considered as future steps for the proposed idea.
\label{sec:typestyle}

% References should be produced using the bibtex program from suitable
% BiBTeX files (here: strings, refs, manuals). The IEEEbib.bst bibliography
% style file from IEEE produces unsorted bibliography list.
% -------------------------------------------------------------------------
\bibliographystyle{IEEEbib}
\bibliography{strings,refs}

\end{document}